%% file: main.tex
\pdfoutput=1
\documentclass[letterpaper, 10 pt, conference]{ieeeconf}

\IEEEoverridecommandlockouts
\usepackage{amsmath}
\usepackage{amssymb}
\usepackage{bm}
\usepackage{booktabs}
\usepackage{multirow}

\usepackage{graphicx}
\usepackage{subfig}
\usepackage[table]{xcolor}

\newcommand{\Dmax}{D_{\max}}

\newcommand{\Rsrc}{\mathcal{R}}
\newcommand{\Rhat}{\widehat{\mathcal{R}}}

\title{\LARGE \bf
Learning Distance-Conditioned Object Transport for Humanoid
Loco-Manipulation from a Single Motion Clip
}

\author{
Yuhyeon Hwang$^{1}$,
Daniel Sungho Jung$^{2}$,
YongHyeok Seo$^{1,3}$,
Mingi Jung$^{1}$,\\
Chang Nho Cho$^{1}$,
Jung-Hoon Hwang$^{1}$,
and Dongin Shin$^{1}$%
\thanks{This work was supported by the Institute of Information \&
Communications Technology Planning \& Evaluation
(IITP) grant funded by the Korean government
(MSIT) (No. RS-2025-25441838) and
the Korea Electronics Technology Institute (KETI) through its Fundamental
Research Support Program.}%
\thanks{$^{1}$Korea Electronics Technology Institute, Republic of Korea.}%
\thanks{$^{2}$Seoul National University, Republic of Korea.}%
\thanks{$^{3}$Korea University, Republic of Korea.}%
}

\begin{document}

\maketitle
\thispagestyle{empty}
\pagestyle{empty}

%%%%%%%%%%%%%%%%%%%%%%%%%%%%%%%%%%%%%%%%%%%%%%%%%%%%%%%%%%%%%%%%%%%%%%%%%%%%%%%%
\begin{abstract}
Motion tracking can reproduce humanoid loco-manipulation from a single
retargeted motion clip, but a policy trained on a fixed reference primarily
reproduces its demonstrated transport outcome.
Although the source trajectory visits intermediate object displacements,
transport termination is demonstrated only at its endpoint.
We identify this mismatch as the \emph{termination-versus-passage gap}:
intermediate displacements are observed as passage states rather than
termination-complete outcomes.
We introduce \emph{Distance-Conditioned Reference Recomposition} (DCRR),
which relocates the demonstrated termination segment to intermediate
transport states.
A frozen tracking teacher replays the recomposed references under closed-loop
dynamics, and the retained trajectories are relabeled by their achieved object
placements and distilled into a reference-free policy.
This procedure constructs distance-conditioned supervision from the
interaction behavior encoded in the source motion.
Across Carry, Kick-Push, Crouch-Push, and Drag, DCRR-BC produces
command-dependent transport with an overall normalized distance mean absolute error (MAE) of
$0.15$, compared with $0.28$ for source-only behavior cloning.
RL fine-tuning further improves the command response and execution robustness
in the training simulator and under sim-to-sim transfer.
Finally, hardware experiments demonstrate transport-distance modulation
across all four interaction modes.
\end{abstract}

%%%%%%%%%%%%%%%%%%%%%%%%%%%%%%%%%%%%%%%%%%%%%%%%%%%%%%%%%%%%%%%%%%%%%%%%%%%%%%%%
\input{sec_introduction}
\input{sec_related_work}
\input{sec_problem_formulation}
\input{sec_dcrr}
\input{sec_experiments}
\input{sec_conclusion}

\end{document}

%% file: sec_introduction.tex
\section{INTRODUCTION}

Learning from human motion references has enabled humanoid robots to acquire
coordinated whole-body behaviors through explicit reference tracking and
learned motion priors \cite{peng2018deepmimic,peng2021amp}.
Recent methods have scaled these paradigms to large motion collections
\cite{he2024omnih2o,fu2024humanplus}, versatile skill composition
\cite{liao2025beyondmimic}, and whole-body object interaction
\cite{hassan2023interphys,wang2025physhsi}.
However, a policy trained with a fixed motion reference primarily reproduces
the demonstrated task outcome.
In object transport, a single clip may demonstrate how to approach, interact
with, transport, and release an object, but provides complete termination
behavior at only one transport distance.
We study distance-conditioned humanoid object transport from a single source
motion clip for each interaction mode.
Given a distance command, the policy must transport the object by the
requested distance and terminate the interaction near the corresponding goal,
rather than reproduce the source endpoint.
We consider four interaction modes---Carry, Kick-Push, Crouch-Push, and
Drag---that share interaction initiation, active transport, and termination
while requiring different contact strategies.

\begin{figure}[t]
\centering
\includegraphics[width=0.95\linewidth]{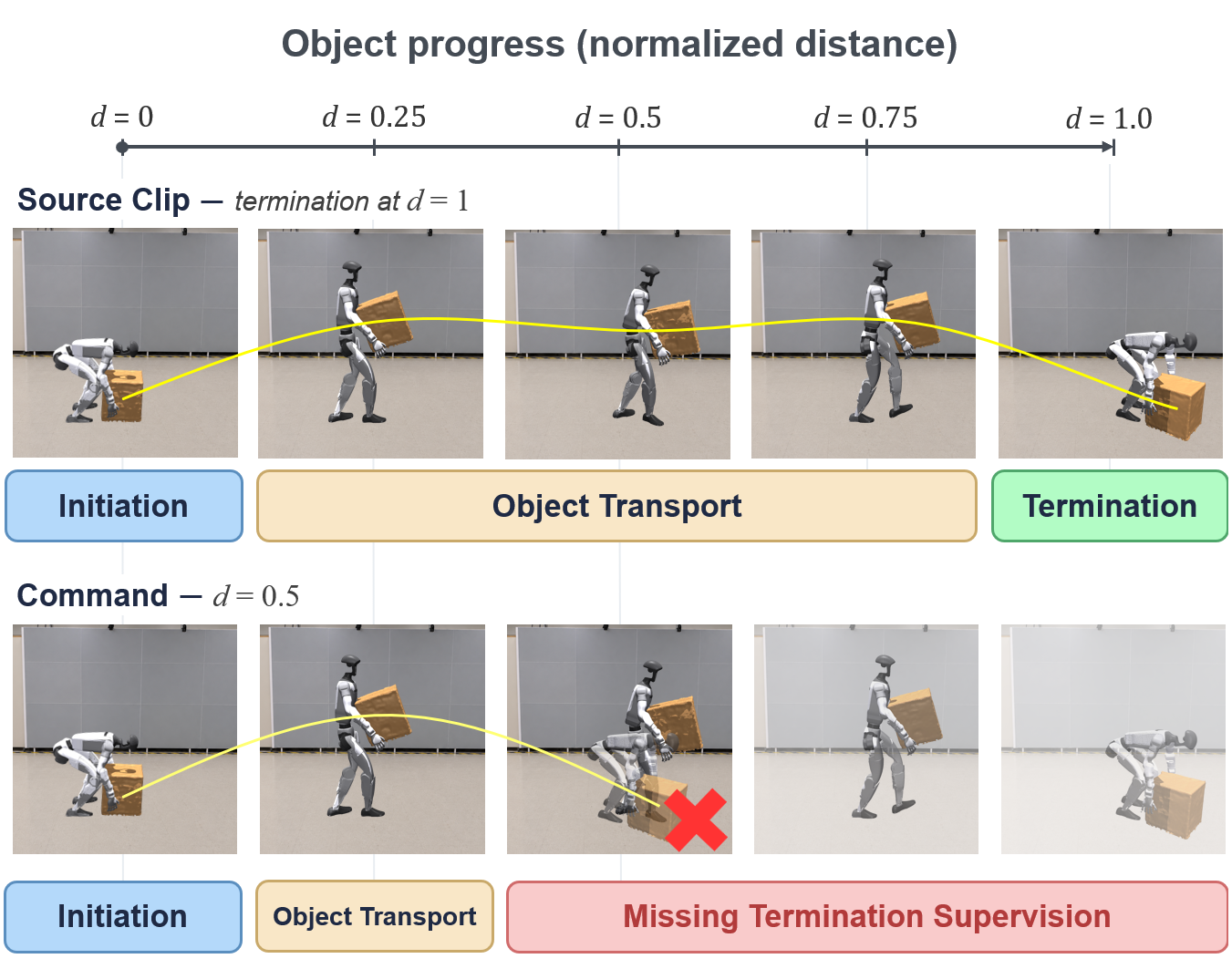}
\caption{The termination-versus-passage gap in a single source motion:
intermediate displacements occur during transport, while termination is
demonstrated only at the source endpoint.}
\label{fig:single-clip}
\end{figure}

The key challenge is the absence of termination supervision at intermediate
distances.
As illustrated in Fig.~\ref{fig:single-clip}, intermediate displacements are
observed while transport continues, whereas termination behavior is
demonstrated only near the source endpoint.
For Carry, termination at an intermediate distance requires lowering and
releasing the object, whereas the source motion continues transporting it
through that displacement.
We refer to this mismatch as the
\emph{termination-versus-passage gap}.

Providing a distance command does not create the missing termination behavior.
Similarly, hindsight methods such as HER~\cite{andrychowicz2017her} and
GCSL~\cite{ghosh2021gcsl} change the goal assigned to recorded behavior
without changing its actions.
Relabeling an intermediate passage state therefore does not provide the
behavior required to terminate transport at that distance.

We introduce \emph{Distance-Conditioned Reference Recomposition} (DCRR).
DCRR truncates active transport and relocates the demonstrated termination
segment to an intermediate transport state.
Because the transition to the relocated termination segment may not be
dynamically feasible, a frozen tracking teacher replays each recomposed
reference in closed loop.
The retained trajectories are relabeled with their achieved object placements
and distilled into a reference-free policy.
This procedure constructs termination-complete supervision for multiple
transport distances without additional motion demonstrations.

Our contributions are:
(i) a formulation of single-source distance-conditioned transport that
identifies the termination-versus-passage gap;
(ii) DCRR, which constructs termination-complete demonstrations through
reference recomposition and closed-loop replay, relabels them by their
achieved outcomes, and distills them into a reference-free policy; and
(iii) an evaluation across four interaction modes showing command-dependent
transport after distillation, improved execution robustness through RL
fine-tuning, and hardware transfer.
%%%%%%%%%%%%%%%%%%%%%%%%%%%%%%%%%%%%%%%%%%%%%%%%%%%%%%%%%%%%%%%%%%%%%%%%%%%%%%%%

%% file: sec_related_work.tex
%%%%%%%%%%%%%%%%%%%%%%%%%%%%%%%%%%%%%%%%%%%%%%%%%%%%%%%%%%%%%%%%%%%%%%%%%%%%%%%%
\section{RELATED WORK}

\subsection{Motion Tracking and Interaction Priors}

Physics-based imitation learns coordinated whole-body behavior through
explicit reference tracking or learned motion priors.
DeepMimic~\cite{peng2018deepmimic} represents the former, whereas
AMP~\cite{peng2021amp} replaces explicit tracking with an adversarial motion
prior.
Subsequent humanoid methods extend these paradigms to whole-body
teleoperation~\cite{he2024omnih2o,fu2024humanplus} and versatile skill
composition~\cite{liao2025beyondmimic}.
PHP~\cite{wu2026php} combines PPO~\cite{schulman2017ppo} with an imitation
loss maintained throughout policy distillation, an approach related to our
use of frozen-BC and demonstration regularization during RL fine-tuning.
AdaMimic~\cite{huang2025adamimic} adapts a single reference to continuous
task variables through sparse keyframe editing and learned time warping, but
does not model object transport or contact-dependent termination.

Humanoid object interaction has been addressed through physics-based
imitation that incorporates object states~\cite{hassan2023interphys,wang2025physhsi}, residual or hierarchical
control~\cite{zhao2025resmimic,yin2025visualmimic}, and reference-free
distillation of video-derived priors~\cite{wu2026sugar}.
InterReal~\cite{liang2026interreal} augments a source interaction while
preserving hand--object contacts, but retains its transport outcome.
Reward-based approaches can instead learn goal-directed box transport without
a motion prior~\cite{zhang2024wococo}.
DCRR focuses on constructing different termination-complete transport
outcomes from one source interaction.

\subsection{Reference Transformation and Demonstration Generation}

Classical motion editing deforms recorded motions under spatial or temporal
constraints~\cite{witkin1995warp,gleicher2001path}, while motion graphs
recompose segments through transitions between compatible
frames~\cite{kovar2002motion}.
Physics-based control can also derive parameterized behavior families from a
single clip~\cite{leeSingleClipFamily}, but does not specifically address
contact-dependent object-transport termination.

Manipulation data-generation methods transform object-relative motion
segments and either replay them in new task configurations
~\cite{mandlekar2023mimicgen,garrett2024skillmimicgen} or synthesize
trajectories offline~\cite{xue2025demogen}.
For humanoid loco-manipulation, DemoHLM~\cite{fu2026demohlm} generates
spatial variations from one demonstration using object-relative motion
representations.
Humanoid-DART~\cite{debbad2026humanoiddart} instead expands sparse
demonstrations through diffusion-based generation, physical evaluation, and
hindsight relabeling.
DCRR targets terminal-outcome variation by relocating the source clip's own
termination behavior, without a learned generator or additional
demonstrations, and validates the recomposed references through closed-loop
replay.

OmniRetarget~\cite{yang2025omniretarget} produces humanoid references while
preserving robot--object--terrain relationships under embodiment and scene
variations.
Given a retargeted source reference, DCRR instead focuses on varying its
terminal transport outcome.

\subsection{Goal-Conditioned Imitation and Relabeling}

HER~\cite{andrychowicz2017her} and GCSL~\cite{ghosh2021gcsl} expand
goal-conditioned supervision by assigning achieved outcomes as alternative
goals.
Relabeling, however, changes the assigned goal without changing the recorded
action.
Prior analyses show that such actions can be suboptimal for the relabeled goal
and that supervised goal-conditioned policies can struggle with combinations
not represented in the original trajectories
~\cite{zhang2022hindsight,ghugare2024stitching}.

For single-source distance-conditioned transport, intermediate object
displacements appear as passage states followed by continued transport.
Relabeling a passage state with an intermediate goal changes the assigned
goal but does not create the termination behavior required to stop there.
This motivates constructing termination-complete demonstrations for
intermediate outcomes.
%%%%%%%%%%%%%%%%%%%%%%%%%%%%%%%%%%%%%%%%%%%%%%%%%%%%%%%%%%%%%%%%%%%%%%%%%%%%%%%%

%% file: sec_problem_formulation.tex
\section{PROBLEM FORMULATION}
\label{sec:problem}

\subsection{Distance-Conditioned Object Transport}
\label{sec:problem-task}

We consider a single retargeted source reference
$\Rsrc=(r_0,\ldots,r_T)$
that demonstrates one complete humanoid object-transport interaction.
Each frame contains the reference robot and object states.
The interaction begins from a stable pre-interaction state and ends after
the robot and object reach a stable task-completion state.

Let $p_{\mathrm{obj}}(t)\in\mathbb R^3$ denote the object position, and let
$\bar p_{\mathrm{obj}}(t)\in\mathbb R^2$ denote its planar component.
We define the source transport direction $u$ and transport-distance scale
$\Dmax$ as
\begin{equation}
\begin{aligned}
u &=
\frac{
\bar p_{\mathrm{obj}}(T)-\bar p_{\mathrm{obj}}(0)
}{
\left\|
\bar p_{\mathrm{obj}}(T)-\bar p_{\mathrm{obj}}(0)
\right\|_2
},\\
\Dmax &=
\left\|
\bar p_{\mathrm{obj}}(T)-\bar p_{\mathrm{obj}}(0)
\right\|_2.
\end{aligned}
\label{eq:distance-scale}
\end{equation}
The clip-specific scale $\Dmax$ provides a common normalized command space
across source motions with different metric transport distances.
The planar object displacement magnitude at frame $t$ is
\begin{equation}
D_{\Rsrc}(t)
=
\left\|
\bar p_{\mathrm{obj}}(t)
-
\bar p_{\mathrm{obj}}(0)
\right\|_2.
\label{eq:progress}
\end{equation}
For the source motions considered, this displacement is approximately
monotonic during active transport and is used to identify transport states
for reference recomposition.

Given a desired transport distance $D$, we define the normalized command as
\begin{equation}
d := \frac{D}{\Dmax}.
\label{eq:normalized-command}
\end{equation}
The corresponding planar goal is positioned along the source transport
direction:
\begin{equation}
\bar g(d)
=
\bar p_{\mathrm{obj}}(0)
+
d\Dmax u.
\label{eq:goal}
\end{equation}

Let $g\in\mathbb R^3$ denote the three-dimensional object goal.
For a nominal distance command $d$, the planar component of $g$ is
$\bar g(d)$, while its vertical component is inherited from the terminal
object state of the source reference.
Thus, $d=0$ corresponds to the initial object position, $d=1$ to the source endpoint, and $d>1$ to an extrapolated planar goal.
The policy is conditioned on the relative object-to-goal vector
$g-p_{\mathrm{obj}}(t)$; the scalar command $d$ is not provided separately.

Let $T_{\mathrm{ep}}$ denote the end of a policy rollout.
We measure the achieved normalized transport distance as
\begin{equation}
\hat d
=
\frac{
\left\|
\bar p_{\mathrm{obj}}(T_{\mathrm{ep}})
-
\bar p_{\mathrm{obj}}(0)
\right\|_2
}{
\Dmax
}.
\label{eq:achieved-distance}
\end{equation}
The task objective is to reach a stable completion state while minimizing the
distance error $|\hat d-d|$.

\subsection{Termination-versus-Passage Gap}
\label{sec:term-vs-passage}

Although the source reference visits intermediate object displacements on
its way to $d=1$, these displacements are observed during ongoing transport
and do not generally represent completed outcomes.
Only near the source endpoint does the reference demonstrate the
trajectory-level transition of the robot and object from active transport to
task completion.
Depending on the interaction mode, this transition may involve behaviors such as object
placement and release, contact disengagement, or post-contact recovery.
An intermediate object displacement therefore provides evidence of passage,
but not the behavior required to terminate transport at that distance.

Goal relabeling can associate a passage state with an intermediate command,
but it does not change the recorded actions or create the missing termination
behavior.
We refer to this mismatch as the
\emph{termination-versus-passage gap}.
DCRR addresses this gap by recomposing the demonstrated termination behavior
at intermediate transport states and validating the resulting references
through closed-loop replay.

%% file: sec_dcrr.tex
%%%%%%%%%%%%%%%%%%%%%%%%%%%%%%%%%%%%%%%%%%%%%%%%%%%%%%%%%%%%%%%%%%%%%%%%%%%%%%%%
\section{DISTANCE-CONDITIONED REFERENCE RECOMPOSITION}
\label{sec:method}

\begin{figure}[t]
\centering
\includegraphics[width=0.75\linewidth]{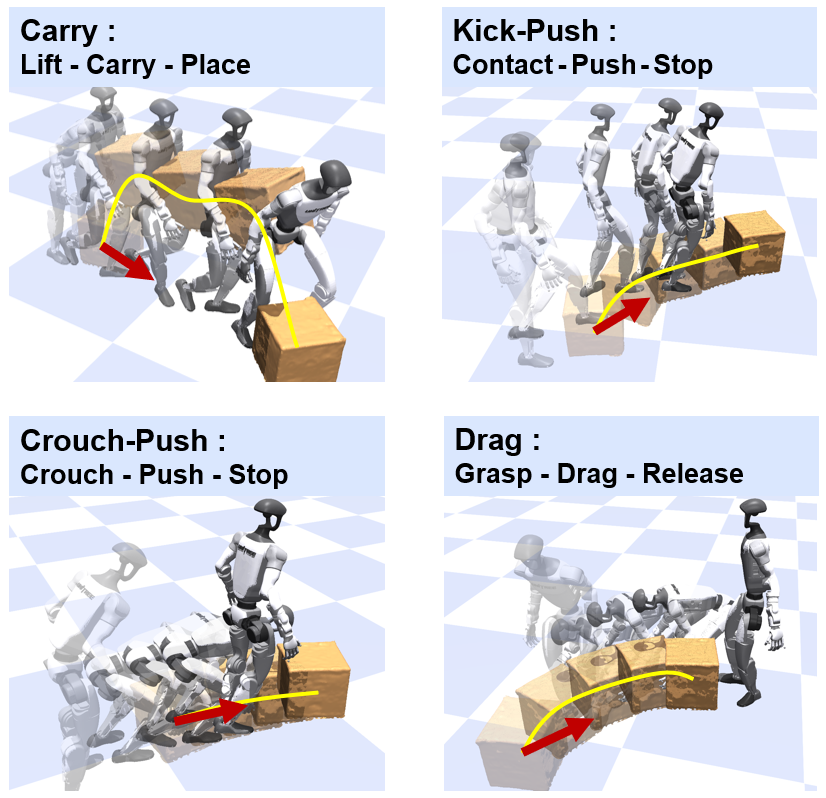}
\caption{Four source motions for object transport: Carry, Kick-Push,
Crouch-Push, and Drag.}
\label{fig:source-motions}
\end{figure}

\begin{figure*}[!t]
\centering
\includegraphics[width=0.96\textwidth]{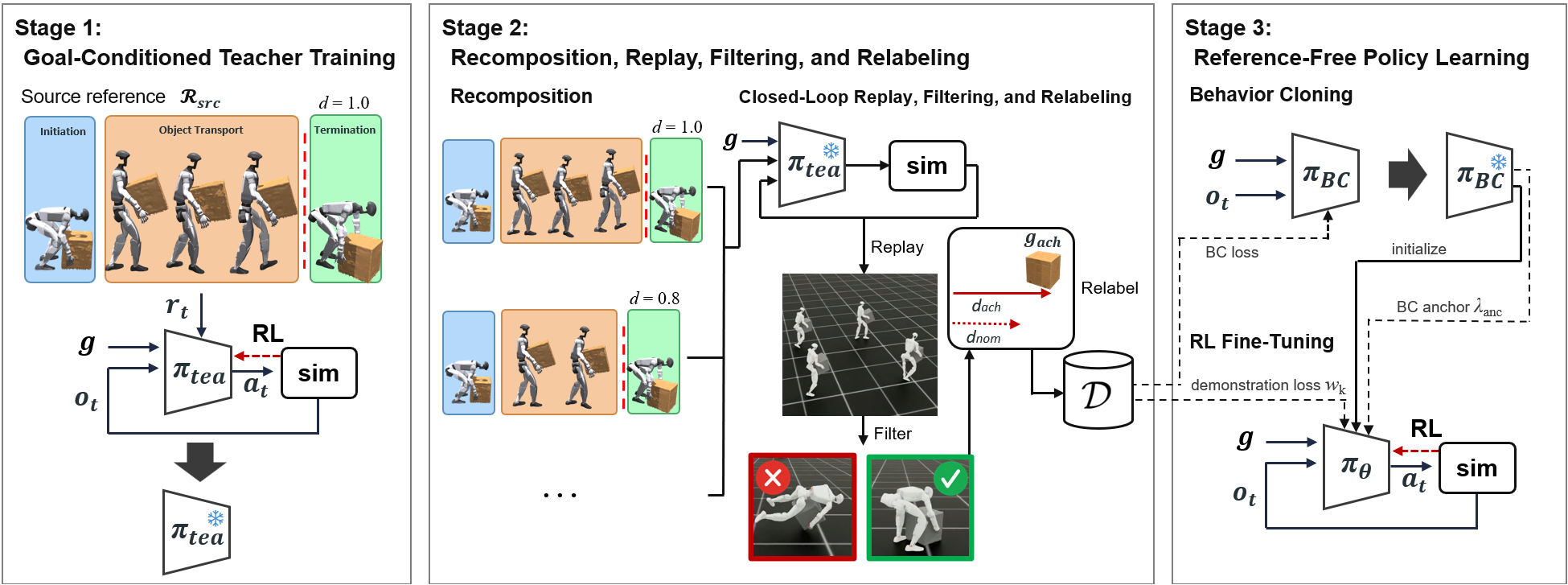}
\caption{Three-stage DCRR pipeline: goal-conditioned teacher training;
reference recomposition, closed-loop replay, filtering, and achieved-goal
relabeling; reference-free policy learning through distillation and
RL fine-tuning.}
\label{fig:pipeline}
\end{figure*}

Distance-Conditioned Reference Recomposition (DCRR) constructs
distance-conditioned supervision from a single source reference in three
stages.
First, a goal-conditioned tracking teacher is trained on the source
reference.
Second, the source motion is recomposed to produce candidate terminal
outcomes, which are replayed, filtered, and relabeled using the frozen
teacher.
Third, the retained trajectories are distilled into a reference-free policy
and further fine-tuned through reinforcement learning.
Fig.~\ref{fig:source-motions} shows the four source interactions considered
in this work---Carry, Kick-Push, Crouch-Push, and Drag---which involve
different contact and object-transport mechanisms.
Fig.~\ref{fig:pipeline} illustrates the three-stage DCRR pipeline.

%%%%%%%%%%%%%%%%%%%%%%%%%%%%%%%%%%%%%%%%%%%%%%%%%%%%%%%%%%%%%%%%%%%%%%%%%%%%%%%%
\subsection{Goal-Conditioned Tracking Teacher}
\label{sec:method-teacher}

For each source motion, we train a goal-conditioned tracking teacher
$\pi_{\mathrm{tea}}$ using PPO~\cite{schulman2017ppo}.
We adopt the robot- and object-tracking formulation of
OmniRetarget~\cite{yang2025omniretarget} and augment it with a goal-reaching term.
Its observation includes robot proprioception, phase-indexed reference joint
positions and velocities, reference orientation relative to the robot base,
and an object-to-goal observation.
The planar goal corresponds to $\bar g(d)$ defined in
Sec.~\ref{sec:problem-task}; the scalar command $d$ is not provided
separately.

To improve robustness during closed-loop replay, we perturb the terminal
object goal along the source transport direction and its planar orthogonal
direction.
After training, the frozen teacher is used to replay both the source and
recomposed references without candidate-specific adaptation.
%%%%%%%%%%%%%%%%%%%%%%%%%%%%%%%%%%%%%%%%%%%%%%%%%%%%%%%%%%%%%%%%%%%%%%%%%%%%%%%%
\subsection{Recomposition, Replay, and Relabeling}
\label{sec:method-replay}

DCRR constructs candidate references by relocating the demonstrated
transport-termination behavior to different points along the source
interaction.
We detect active transport from smoothed planar object motion and enumerate
feasible splice frames
$t\in\mathcal S_{\Rsrc}$
within the detected transport support.
The same criteria are applied to all source motions without motion-specific
frame annotations.

For each candidate splice $t$, we select a source frame from which to reuse
the terminal portion of the motion.
The residual transport of a tail beginning at frame $f$ is
$c_{\mathrm{term}}(f)
= D_{\Rsrc}(T)-D_{\Rsrc}(f)$.
Candidate tails must retain sufficient duration and have an object height
compatible with that at the splice.
We restrict the candidates to tails whose residual transport is within
$\delta\Dmax$ of the minimum feasible value and select
\begin{equation}
f^\star(t)
=
\operatorname*{arg\,min}_{f\in\mathcal A_\delta(t)}
\rho(t-1,f),
\label{eq:tail-selection}
\end{equation}
where $\mathcal A_\delta(t)$ denotes the resulting candidate set.
The transition cost $\rho$ is evaluated between source frames before planar
alignment and combines robot-state, robot-relative object-pose, and velocity
discrepancies.
The selection then minimizes $\rho$ over candidates with near-minimal
residual transport.

The source reference is truncated immediately before the splice and
concatenated with the selected terminal tail:
\begin{equation}
\Rhat(t)
=
(r_0,\ldots,r_{t-1})
\oplus
\mathrm T_{\Delta(t)}
(r_{f^\star(t)},\ldots,r_T),
\label{eq:recomposition}
\end{equation}
where $\oplus$ denotes temporal concatenation.
The transformation $\mathrm T_{\Delta(t)}$ applies a planar rotation to align
the initial robot heading of the reused tail with that of the final prefix
frame and translates the tail to align its initial object position with the
object position at the splice.
The transformation is applied consistently to the global robot and object
states without temporal blending or motion generation.

We approximate the nominal transport distance of a candidate as
\begin{equation}
D_{\mathrm{nom}}(t)
=
D_{\Rsrc}(t-1)
+
c_{\mathrm{term}}\!\left(f^\star(t)\right),
\label{eq:nominal-distance}
\end{equation}
with normalized value
$d_{\mathrm{nom}}(t)=D_{\mathrm{nom}}(t)/\Dmax$.
For a desired command $d$, we select the candidate by minimizing
the nominal distance error together with the transition cost:
\begin{equation}
t_s(d)
=
\operatorname*{arg\,min}_{t\in\mathcal S_{\Rsrc}}
\left[
\left|
D_{\mathrm{nom}}(t)-d\Dmax
\right|
+
\lambda_s
\rho\!\left(t-1,f^\star(t)\right)
\right],
\label{eq:splice-selection}
\end{equation}
where $\lambda_s$ weights the transition cost.
We use $\delta=0.02$ and $\lambda_s=0.05$ for all source motions.
The selected reference $\Rhat(t_s(d))$ provides a nominal proposal for the
requested transport distance.

Because a splice may occur while the object is moving, a recomposed reference
does not guarantee a valid transition or terminal outcome.
We therefore replay the source and recomposed references using the frozen
tracking teacher under closed-loop robot dynamics and contact interactions.
We retain only replays that avoid robot falls, produce sufficient object
transport, reach a valid terminal object state, and satisfy a source-derived
tracking-error criterion.

Let $g_{\mathrm{ach}}\in\mathbb R^3$ denote the terminal object placement of a
retained replay.
We denote its normalized planar transport distance, computed as in
Eq.~\eqref{eq:achieved-distance}, by $d_{\mathrm{ach}}$.
Each retained trajectory is relabeled with $g_{\mathrm{ach}}$ rather than its
nominal reference goal, without changing the recorded states or actions.
The achieved placement is converted into the corresponding object-to-goal
observations.
%%%%%%%%%%%%%%%%%%%%%%%%%%%%%%%%%%%%%%%%%%%%%%%%%%%%%%%%%%%%%%%%%%%%%%%%%%%%%%%%
\subsection{Reference-Free Policy Learning}
\label{sec:method-distill}

Let $\mathcal D$ denote the retained source and recomposed replay
trajectories after achieved-goal relabeling.
Each action in $\mathcal D$ is the deterministic output of the frozen
tracking teacher during replay.
For $d_a$ action dimensions, we define the dimension-averaged action error as
$\ell_{\mathrm{act}}(x,y)=d_a^{-1}\|x-y\|_2^2$.

We first train a reference-free policy $\pi_{\mathrm{BC}}$ by behavior
cloning:
\begin{equation}
\mathcal L_{\mathrm{BC}}(\theta)
=
\mathbb E_{(o_t,a_t)\sim\mathcal D}
\left[
\ell_{\mathrm{act}}
\left(\pi_\theta(o_t),a_t\right)
\right].
\label{eq:bc}
\end{equation}
The policy's input consists of robot proprioception, observation history,
and object-to-goal observations.

We further refine the distilled policy using PPO with a frozen-BC
action-mean anchor and a DAPG-style demonstration loss
\cite{rajeswaran2018dapg}.
The actor objective is
\begin{equation}
\begin{split}
\mathcal L_{\mathrm{actor}}(\theta)
={}&
\mathcal L_{\mathrm{PPO}}(\theta)
+
\lambda_{\mathrm{anc}}
\mathbb E_{o\sim \nu_{\pi_\theta}}
\left[
\ell_{\mathrm{act}}
\left(\mu_\theta(o),\pi_{\mathrm{BC}}(o)\right)
\right]
\\
&+
w_k
\mathbb E_{(o,a)\sim\mathcal D}
\left[
\ell_{\mathrm{act}}
\left(\mu_\theta(o),a\right)
\right],
\end{split}
\label{eq:rlft}
\end{equation}
where $\mu_\theta$ denotes the current actor's action mean,
$\nu_{\pi_\theta}$ its on-policy observation distribution, and
$\pi_{\mathrm{BC}}(o)$ the action output of the frozen BC policy.
The BC anchor penalizes deviations of the actor's action mean from
the frozen BC policy's output at on-policy observations.
The DAPG-style loss penalizes deviations from teacher actions recorded
during replay and paired with the corresponding reference-free student
observations.
We use $\lambda_{\mathrm{anc}}=10$ and set
$w_k=\max_t |A^{\mathrm{raw}}_{k,t}|$,
where $A^{\mathrm{raw}}_{k,t}$ denotes the unnormalized on-policy
advantage at iteration $k$.
The RL reward combines goal progress and task completion with fall
and regularization penalties and contains no reference-tracking term.
%%%%%%%%%%%%%%%%%%%%%%%%%%%%%%%%%%%%%%%%%%%%%%%%%%%%%%%%%%%%%%%%%%%%%%%%%%%%%%%%

%% file: sec_experiments.tex
%%%%%%%%%%%%%%%%%%%%%%%%%%%%%%%%%%%%%%%%%%%%%%%%%%%%%%%%%%%%%%%%%%%%%%%%%%%%%%%%
\section{EXPERIMENTAL RESULTS AND DISCUSSION}
\label{sec:experiments}

We evaluate DCRR in terms of closed-loop replay validity, distance
conditioning, achieved-goal relabeling, RL fine-tuning, and hardware transfer.

%%%%%%%%%%%%%%%%%%%%%%%%%%%%%%%%%%%%%%%%%%%%%%%%%%%%%%%%%%%%%%%%%%%%%%%%%%%%%%%%
\subsection{Experimental Setup}
\label{sec:exp-setup}

We evaluate four humanoid object-transport interactions: Carry, Kick-Push,
Crouch-Push, and Drag.
Carry requires grasping, lifting, transporting, and releasing the object.
Kick-Push transfers the object through a brief foot impact, Crouch-Push
maintains pushing contact in a lowered posture, and Drag pulls the object
during backward locomotion while maintaining hand--object contact.
For each interaction, we use a single retargeted source motion~\cite{yang2025omniretarget}, with source
transport distances of $1.44$, $1.30$, $0.80$, and $1.10\,\mathrm{m}$,
respectively.
An independent policy is trained for each source motion using the same DCRR
pipeline.

We use Isaac Sim as the training simulator and the Unitree G1 as the robot platform.
Both the BC and RLFT policies use a $512$--$256$--$128$ ELU MLP with a
one-frame joint-state history and output $29$ joint-position targets.
The goal-reaching term used during teacher training and RL fine-tuning is
$r_{\mathrm{goal}}
=2\exp(-\|p_{\mathrm{obj}}-g\|_2^2/0.3^2)$.
During training, we randomize the object mass over
$[0.1,2.0]\,\mathrm{kg}$ and the friction coefficient over $[0.1,1.0]$.
For RL fine-tuning, distance commands are sampled uniformly from
$[0.1,1.2]$.

We report the achieved normalized transport distance $\hat d$ defined in
Eq.~\eqref{eq:achieved-distance}, the distance MAE $|\hat d-d|$, and the
robot fall rate.
Each method--motion--command setting is evaluated over $300$ episodes in
simulation.
Overall values are macro-averaged across the four interaction modes.

\subsection{Closed-Loop Replay of Recomposed References}
\label{sec:exp-recomposition}

We evaluate whether closed-loop replay of recomposed references produces
physically valid intermediate-distance trajectories without motion generation,
trajectory optimization, or teacher adaptation.

Each reference is replayed by a frozen teacher trained on the original
$d=1.0$ source reference.
We compare a \emph{Parallel} teacher trained with goal perturbations of
$\pm30\%$ along the transport direction against a
\emph{Parallel $+$ Orthogonal} variant that additionally perturbs the goal
in the orthogonal direction.
A replay is retained if it completes without falling, transports the object
to a valid terminal state, and satisfies the source-derived tracking-error
threshold.

\begin{table}[!t]
\centering
\caption{%
Achieved distance and closed-loop replay validity of recomposed references,
 across four motions.
}
\label{tab:recomposition-replay}

\footnotesize
\setlength{\tabcolsep}{5pt}
\renewcommand{\arraystretch}{1.1}

\begin{tabular}{@{}lcccc@{}}
\toprule
& \multicolumn{2}{c}{$d_{\mathrm{ach}}$}
& \multicolumn{2}{c}{Valid replay (\%)} \\
\cmidrule(lr){2-3}
\cmidrule(l){4-5}
$d_{\mathrm{nom}}$
& Parallel
& $+$ Orth.
& Parallel
& $+$ Orth. \\
\midrule
0.2
& 0.08
& 0.12
& 28.6
& \textbf{90.7} \\

0.4
& 0.20
& 0.28
& 62.0
& \textbf{98.6} \\

0.6
& 0.43
& 0.48
& 94.6
& \textbf{99.4} \\

0.8
& 0.71
& 0.71
& \textbf{98.6}
& 93.3 \\

1.0 (Source)
& 0.92
& 0.92
& 94.7
& 94.8 \\
\midrule
Overall
& --
& --
& 75.7
& \textbf{95.4} \\
\bottomrule
\end{tabular}
\end{table}

Table~\ref{tab:recomposition-replay} reports the achieved distance
$d_{\mathrm{ach}}$ and valid-replay rate at each nominal target.
Orthogonal perturbations provide the largest gains for short-distance
recompositions: validity increases from 28.6\% to 90.7\% at
$d_{\mathrm{nom}}=0.2$ and from 62.0\% to 98.6\% at
$d_{\mathrm{nom}}=0.4$, raising the overall rate from 75.7\% to 95.4\%.
The two variants are comparable near the source distance, where the
recomposed references deviate less from the original motion.

Replay validity does not, however, guarantee agreement between the nominal
and achieved outcomes.
For example, at $d_{\mathrm{nom}}=0.2$, the orthogonally augmented teacher
achieves $d_{\mathrm{ach}}=0.12$ despite a 90.7\% valid-replay rate.
We therefore compare nominal and achieved outcome labels separately in
Sec.~\ref{sec:exp-relabel}.

\subsection{Distance Conditioning from a Single Source Motion}
\label{sec:exp-distance}
\begin{figure*}[!t]
    \centering
    \includegraphics[width=0.97\textwidth]{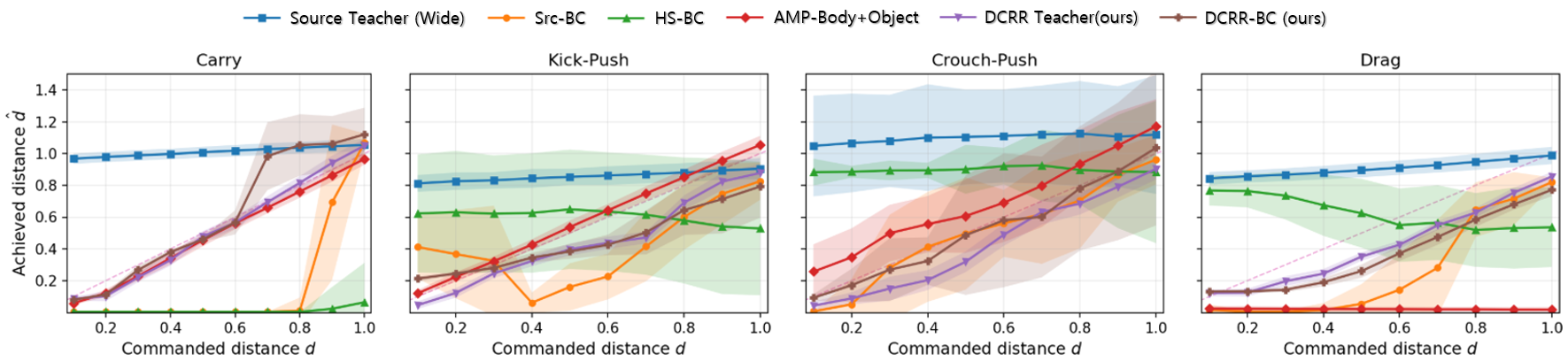}
    \caption{
Command-response curves across four interaction modes.
Markers and shaded regions show the mean and one standard deviation, respectively; the diagonal denotes exact command-response correspondence
($\hat d=d$).
}
    \label{fig:dial}
\end{figure*}

We evaluate whether direct goal conditioning or hindsight relabeling can
bridge the termination-versus-passage gap and what is gained by providing
termination-complete supervision.
Figure~\ref{fig:dial} and Table~\ref{tab:main} compare six methods.

\begin{table*}[!t]
\centering
\caption{
Distance MAE and fall rate across four interaction modes.
Overall values are macro-averaged across interaction modes.
}
\label{tab:main}
\footnotesize
\setlength{\tabcolsep}{3.5pt}
\resizebox{\textwidth}{!}{%
\begin{tabular}{@{}lcc cc cc cc cc@{}}
\toprule
Method
& \multicolumn{2}{c}{Carry}
& \multicolumn{2}{c}{Kick-Push}
& \multicolumn{2}{c}{Crouch-Push}
& \multicolumn{2}{c}{Drag}
& \multicolumn{2}{c}{Overall} \\
\cmidrule(lr){2-3}
\cmidrule(lr){4-5}
\cmidrule(lr){6-7}
\cmidrule(lr){8-9}
\cmidrule(lr){10-11}
& MAE $\downarrow$ & Fall (\%) $\downarrow$
& MAE $\downarrow$ & Fall (\%) $\downarrow$
& MAE $\downarrow$ & Fall (\%) $\downarrow$
& MAE $\downarrow$ & Fall (\%) $\downarrow$
& MAE $\downarrow$ & Fall (\%) $\downarrow$ \\
\midrule
Source Teacher (Wide)
& $0.46{\pm}0.26$ & $0.03$
& $0.33{\pm}0.23$ & $0.07$
& $0.56{\pm}0.39$ & $\mathbf{0.00}$
& $0.37{\pm}0.24$ & $\mathbf{0.00}$
& $0.43$ & $0.03$ \\

Src-BC
& $0.40{\pm}0.26$ & $0.60$
& $0.28{\pm}0.18$ & $29.77$
& $0.16{\pm}0.21$ & $12.70$
& $0.28{\pm}0.20$ & $14.30$
& $0.28$ & $14.34$ \\

HS-BC
& $0.54{\pm}0.29$ & $0.03$
& $0.42{\pm}0.28$ & $5.73$
& $0.41{\pm}0.28$ & $1.27$
& $0.38{\pm}0.22$ & $53.63$
& $0.44$ & $15.17$ \\

AMP-Body+Object
& $\mathbf{0.05{\pm}0.03}$ & $0.30$
& $\mathbf{0.04{\pm}0.04}$ & $0.03$
& $0.20{\pm}0.15$ & $\mathbf{0.00}$
& $0.53{\pm}0.29$ & $\mathbf{0.00}$
& $0.21$ & $0.08$ \\

DCRR Teacher
& $\mathbf{0.05{\pm}0.04}$ & $\mathbf{0.00}$
& $0.11{\pm}0.06$ & $\mathbf{0.00}$
& $\mathbf{0.12{\pm}0.07}$ & $\mathbf{0.00}$
& $\mathbf{0.13{\pm}0.05}$ & $0.03$
& $\mathbf{0.10}$ & $\mathbf{0.01}$ \\

DCRR-BC
& $0.11{\pm}0.15$ & $8.50$
& $0.13{\pm}0.13$ & $6.47$
& $0.18{\pm}0.29$ & $13.83$
& $0.18{\pm}0.08$ & $7.47$
& $0.15$ & $9.07$ \\
\bottomrule
\end{tabular}%
}
\end{table*}

The \textbf{Source Teacher (Wide)} tracks the fixed $d=1.0$ source reference
while being trained with goals spanning $d\in[0.1,1.0]$, where ``Wide''
denotes the training goal distribution.
All replay-based variants use the same frozen source teacher trained on the
original $d=1.0$ reference with goal perturbations in the parallel and
orthogonal directions.
\textbf{Src-BC} distills source-reference replays into a reference-free
policy, while \textbf{HS-BC} additionally relabels source-trajectory prefixes
with their intermediate outcomes without providing corresponding termination
behavior.
\textbf{DCRR Teacher} replays command-specific recomposed references, and
\textbf{DCRR-BC} distills the resulting replays.
\textbf{AMP-Body+Object} provides an additional reference-free baseline using
an adversarial prior over robot and object motion.
Source Teacher (Wide) and DCRR Teacher use reference inputs, whereas the
remaining methods are evaluated without reference inputs.

We evaluate ten commands from $d=0.1$ to $1.0$.
Despite training over the full command range, Source Teacher (Wide) remains
close to the demonstrated endpoint, producing MAEs of $0.33$--$0.56$ across
the four motions.
This indicates that broad goal exposure with a fixed source reference does
not consistently produce intermediate-distance termination.

Src-BC produces command-dependent responses for some motions, but remains
inconsistent across motions and shows substantial fall rates on Crouch-Push,
Drag, and Kick-Push.
HS-BC also fails to produce a consistent command-response relation: its
response is nearly command invariant on Crouch-Push and Kick-Push, collapses
toward short-distance outcomes on Carry, and reaches a $53.63\%$ fall rate on
Drag.
Relabeling intermediate passage states changes their goal labels but does not
provide the release, contact disengagement, or settling behavior required to
terminate transport at those outcomes.
These results are consistent with the termination-versus-passage gap.

AMP-Body+Object achieves low MAEs on Carry and Kick-Push ($0.05$ and $0.04$)
but does not perform consistently across interaction modes.
On Drag, the object remains near its initial position over most commands,
resulting in an MAE of $0.53$.
Moreover, the lifting behavior does not emerge on Carry despite its low
distance error.
The policy transports the object without consistently reproducing the
demonstrated grasp--lift--transport sequence, showing that low distance error
alone does not necessarily reflect preservation of the source interaction
mode.

In contrast, DCRR Teacher produces command-dependent responses across all four
motions, with an overall MAE of $0.10$ and a fall rate of $0.01\%$.
DCRR-BC retains this command-dependent behavior without reference inputs, achieving an
overall MAE of $0.15$, compared with $0.21$ for AMP-Body+Object, $0.28$ for
Src-BC, and $0.44$ for HS-BC.
After distillation, DCRR-BC shows increased rollout variance and an overall
fall rate of $9.07\%$.
Despite these distinct interaction dynamics, DCRR produces command-dependent
transport across all four interaction modes.

\subsection{Effect of Achieved-Goal Relabeling}
\label{sec:exp-relabel}

Because of contact interactions, balance recovery, and residual object
motion, closed-loop replay does not always terminate at the nominal
outcome of a recomposed reference.

\begin{table}[!t]
\centering
\caption{%
Distance MAE under nominal and achieved goal labels.
Both variants use identical replay trajectories and differ only in their
goal labels.
}
\label{tab:relabel-results}

\footnotesize
\setlength{\tabcolsep}{6pt}
\renewcommand{\arraystretch}{1.08}

\begin{tabular}{@{}lcc@{}}
\toprule
& \multicolumn{2}{c}{Goal label} \\
\cmidrule(l){2-3}
Command $d$
& Nominal
& Achieved (ours) \\
\midrule
0.2
& $0.09{\pm}0.02$
& $0.10{\pm}0.03$ \\

0.4
& $0.16{\pm}0.06$
& $0.13{\pm}0.07$ \\

0.6
& $0.19{\pm}0.07$
& $0.20{\pm}0.10$ \\

0.8
& $0.25{\pm}0.06$
& $0.21{\pm}0.11$ \\

1.0
& $0.18{\pm}0.06$
& $0.18{\pm}0.04$ \\
\midrule
Overall MAE $\downarrow$
& $0.17$
& $0.16$ \\

\bottomrule
\end{tabular}
\end{table}

We compare two BC variants that use the same filtered trajectories and differ
only in their goal labels.
\textbf{Nominal-label BC} uses the nominal transport outcome as the goal
label for each replay, whereas \textbf{Achieved-label BC} uses the terminal
object placement observed at the end of closed-loop replay.

Table~\ref{tab:relabel-results} shows that achieved-goal labeling reduces the
overall distance MAE from $0.17$ to $0.16$.
The largest improvements occur at $d=0.4$ and $d=0.8$, where the MAE decreases
from $0.16$ to $0.13$ and from $0.25$ to $0.21$, respectively.
The effect is not uniform: errors are slightly higher at $d=0.2$ and $d=0.6$,
while the two variants perform similarly at $d=1.0$.

\begin{table*}[!t]
\centering
\caption{%
Command response and fall rates in the training simulator and under
sim-to-sim transfer.
Reported $\hat d$ values are mean$\pm$standard deviation over non-fall
episodes.
Values are averaged across four interaction modes.
}
\label{tab:rlft}
\scriptsize
\setlength{\tabcolsep}{2.5pt}
\renewcommand{\arraystretch}{1.1}
\resizebox{\textwidth}{!}{%
\begin{tabular}{@{}lccccccc ccccccc@{}}
\toprule
& \multicolumn{7}{c}{Training Simulator}
& \multicolumn{7}{c}{Sim-to-Sim} \\
\cmidrule(lr){2-8}
\cmidrule(l){9-15}
Method
& $d=0.2$
& $d=0.4$
& $d=0.6$
& $d=0.8$
& $d=1.0$
& $d=1.2$
& Fall (\%) $\downarrow$
& $d=0.2$
& $d=0.4$
& $d=0.6$
& $d=0.8$
& $d=1.0$
& $d=1.2$
& Fall (\%) $\downarrow$ \\
\midrule
DCRR Teacher
& $0.11{\pm}0.03$
& $0.28{\pm}0.04$
& $0.49{\pm}0.05$
& $0.71{\pm}0.05$
& $0.93{\pm}0.05$
& $1.04{\pm}0.06$
& $0.17$
& $0.13{\pm}0.05$
& $0.27{\pm}0.05$
& $0.51{\pm}0.06$
& $0.71{\pm}0.09$
& $0.94{\pm}0.11$
& $1.07{\pm}0.12$
& $0.17$ \\
DCRR-BC
& $0.15{\pm}0.03$
& $0.31{\pm}0.09$
& $0.50{\pm}0.11$
& $0.73{\pm}0.16$
& $0.90{\pm}0.18$
& $1.08{\pm}0.23$
& $8.83$
& $0.13{\pm}0.07$
& $0.21{\pm}0.11$
& $0.41{\pm}0.18$
& $0.60{\pm}0.23$
& $0.80{\pm}0.28$
& $0.96{\pm}0.32$
& $17.78$ \\
DCRR-RLFT
& $0.15{\pm}0.04$
& $0.33{\pm}0.03$
& $0.53{\pm}0.06$
& $0.76{\pm}0.08$
& $0.95{\pm}0.07$
& $1.12{\pm}0.08$
& $2.22$
& $0.13{\pm}0.06$
& $0.32{\pm}0.09$
& $0.50{\pm}0.11$
& $0.71{\pm}0.13$
& $0.92{\pm}0.11$
& $1.10{\pm}0.11$
& $8.50$ \\
\bottomrule
\end{tabular}%
}
\end{table*}

\subsection{Reinforcement-Learning Fine-Tuning}
\label{sec:exp-rlft}

We evaluate whether the RL fine-tuning described in
Sec.~\ref{sec:method-distill} improves the command response and execution
robustness of DCRR-BC.
We compare DCRR-BC with DCRR-RLFT, both of which operate without
reference inputs, and include the reference-conditioned DCRR Teacher
for comparison.
The training-simulator evaluation uses Isaac Sim, while sim-to-sim
transfer is evaluated in MuJoCo without policy adaptation using a
$1.0\,\mathrm{kg}$ object.
We evaluate commands at
$d\in\{0.2,0.4,0.6,0.8,1.0,1.2\}$.
For DCRR Teacher, the original $d=1.0$ source reference is used at
$d=1.2$, which is within the goal-perturbation range used during
teacher training.

Table~\ref{tab:rlft} summarizes the achieved distance and fall rate in both
evaluation settings.
In the training simulator, DCRR-RLFT produces mean responses closer to the
commands than DCRR-BC at most evaluated distances, while reducing the
fall rate from $8.83\%$ to $2.22\%$.
It also reduces response variability, particularly at medium and long
distances.

Under sim-to-sim transfer, DCRR-RLFT similarly improves the command response
over DCRR-BC and reduces the fall rate from $17.78\%$ to $8.50\%$.
The reference-conditioned DCRR Teacher shows a fall rate of $0.17\%$ in
both settings.
Overall, RL fine-tuning improves the command response and execution robustness
of the distilled reference-free policy, although a robustness gap remains
relative to the teacher.

\subsection{Sim-to-Real Evaluation}
\label{sec:exp-real}

\begin{table*}[!t]
\centering
\caption{
Hardware command response of DCRR-RLFT across four interaction modes.
Achieved distances are summarized as mean$\pm$standard deviation over
non-fall trials.
}
\label{tab:hardware}

\scriptsize
\setlength{\tabcolsep}{4pt}
\renewcommand{\arraystretch}{1.05}

\begin{tabular}{@{}lcc cc cc cc@{}}
\toprule
&
\multicolumn{2}{c}{$d=0.4$}
&
\multicolumn{2}{c}{$d=0.7$}
&
\multicolumn{2}{c}{$d=1.0$}
&
\multicolumn{2}{c}{$d=1.2$}
\\
\cmidrule(lr){2-3}
\cmidrule(lr){4-5}
\cmidrule(lr){6-7}
\cmidrule(lr){8-9}

Motion
& $\hat{d}$ & Fall
& $\hat{d}$ & Fall
& $\hat{d}$ & Fall
& $\hat{d}$ & Fall
\\
\midrule

Carry
& $0.36{\pm}0.06$ & 0/5
& $0.67{\pm}0.05$ & 0/5
& $0.96{\pm}0.07$ & 1/5
& $1.09{\pm}0.08$ & 1/5 \\

Kick-Push
& $0.22{\pm}0.13$ & 0/5
& $0.54{\pm}0.16$ & 1/5
& $0.72{\pm}0.17$ & 1/5
& $0.86{\pm}0.16$ & 2/5 \\

Crouch-Push
& $0.33{\pm}0.09$ & 0/5
& $0.60{\pm}0.09$ & 0/5
& $0.91{\pm}0.11$ & 1/5
& $1.07{\pm}0.12$ & 1/5 \\

Drag
& $0.29{\pm}0.14$ & 0/5
& $0.60{\pm}0.13$ & 0/5
& $1.01{\pm}0.09$ & 0/5
& $1.16{\pm}0.10$ & 1/5 \\

\bottomrule
\end{tabular}

\end{table*}

\begin{figure*}[!t]
\centering

\includegraphics[width=0.87\linewidth]{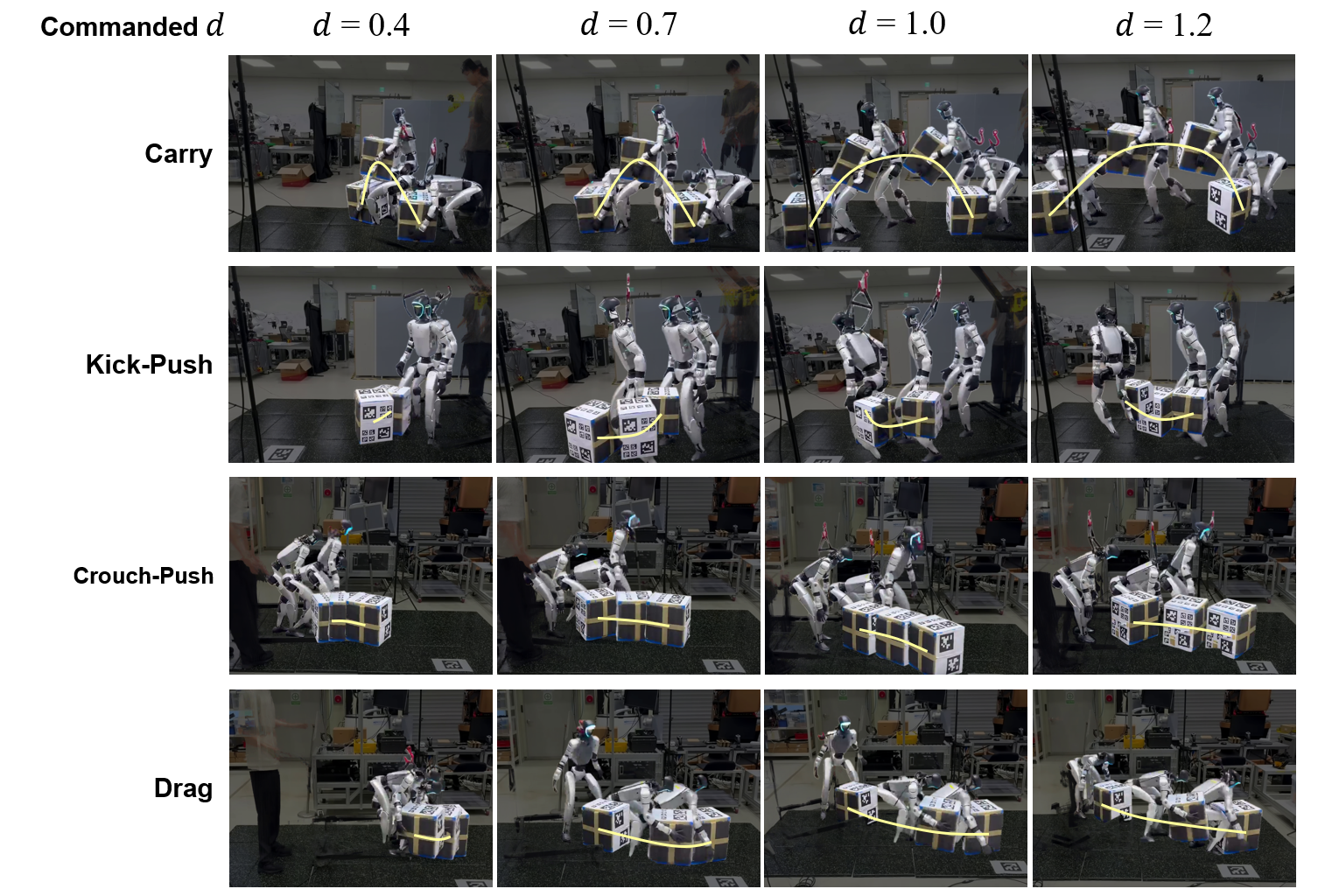}

\caption{
Hardware rollouts of DCRR-RLFT across interaction modes and transport commands. 
}

\label{fig:real}

\end{figure*}

Finally, we evaluate DCRR-RLFT on hardware using a
$1.0\,\mathrm{kg}$ object.
The object pose is estimated using third-person cameras and a visual marker
attached to the object.
At the beginning of each episode, we use the robot's egocentric camera to
align the transport direction with the initial object pose and construct the
commanded goal.
Each interaction mode is evaluated at
$d\in\{0.4,0.7,1.0,1.2\}$ using five trials per command.

Table~\ref{tab:hardware} shows the hardware command response, and
Fig.~\ref{fig:real} shows rollouts across interaction modes and transport
commands.
For each interaction mode, the mean achieved distance increases with the
command.
Across interaction modes, the achieved distances are $0.30$,
$0.60$, $0.90$, and $1.04$ for commands $d=0.4$, $0.7$, $1.0$, and $1.2$,
respectively.

Carry shows relatively small deviations between commanded and achieved
distances, while Crouch-Push and Drag produce distinct responses across the
evaluated commands.
Kick-Push shows larger deviations, consistent with its impulse-driven
interaction.
Falls occurred more frequently at larger commands in these trials,
particularly at $d=1.0$ and $d=1.2$.

%% file: sec_conclusion.tex
%%%%%%%%%%%%%%%%%%%%%%%%%%%%%%%%%%%%%%%%%%%%%%%%%%%%%%%%%%%%%%%%%%%%%%%%%%%%%%%%
\section{CONCLUSION AND LIMITATIONS}

We studied distance-conditioned humanoid object transport from a single source
motion.
DCRR constructs intermediate-distance termination supervision by relocating
the demonstrated termination segment, replaying the recomposed references with
a frozen teacher, relabeling the retained trajectories by their achieved
outcomes, and distilling them into a reference-free policy.
Across four interaction modes, DCRR produces more consistent command-dependent
behavior than direct goal conditioning or relabeling.
RL fine-tuning further improves command-response accuracy and execution
robustness, and the resulting policies produce command-dependent transport on hardware.

DCRR remains limited by the temporal and spatial coverage of its source
motion.
Because it reuses observed motion segments, supporting transport substantially
beyond the source distance may require extending or generating
phase-consistent active transport behavior.
Although the policy uses a three-dimensional goal representation, its training
goals remain concentrated around the source-defined direction, with limited
coverage of other directions.
The reference-conditioned teacher also shows lower fall rates than the
reference-free policy, indicating a remaining robustness gap.
Future work should expand the spatial and temporal coverage and improve reference-free robustness
under contact and dynamics variation.

%%%%%%%%%%%%%%%%%%%%%%%%%%%%%%%%%%%%%%%%%%%%%%%%%%%%%%%%%%%%%%%%%%%%%%%%%%%%%%%%